\documentclass[11pt]{article}
\usepackage{coling2018}
\usepackage{times}
\usepackage{url}
\usepackage{latexsym}
\usepackage{graphicx}
\usepackage{amsmath}
\usepackage{mathtools}
\usepackage{multirow}
\graphicspath{{Figures/}}
\usepackage{tabularx,tabulary}
\usepackage{subcaption}
\usepackage{url}
\usepackage{amssymb}
\usepackage[linesnumbered,ruled]{algorithm2e}
\usepackage{diagbox}
\usepackage{xcolor}
\usepackage{adjustbox}
\usepackage{epstopdf}

\newcommand*{\affaddr}[1]{#1} 
\newcommand*{\affmark}[1][*]{\textsuperscript{#1}}
\newcommand*{\email}[1]{\texttt{#1}}

\title{Adversarial Domain Adaptation for Variational Neural Language Generation in Dialogue Systems}
\author{%
Van-Khanh Tran\affmark[1,2] and Le-Minh Nguyen\affmark[1]\\
\affaddr{\affmark[1]Japan Advanced Institute of Science and Technology, JAIST\\
					1-1 Asahidai, Nomi, Ishikawa, 923-1292, Japan}\\
\email{\{tvkhanh, nguyenml\}@jaist.ac.jp}\\
\affaddr{\affmark[2]University of Information and Communication Technology, ICTU\\
	Thai Nguyen University, Vietnam}\\
\email{tvkhanh@ictu.edu.vn}
}

\date{}

\begin{document}
\maketitle
\begin{abstract}
Domain Adaptation arises when we aim at learning from source domain a model that can perform acceptably well on a different target domain. It is especially crucial for Natural Language Generation (NLG) in Spoken Dialogue Systems when there are sufficient annotated data in the source domain, but there is a limited labeled data in the target domain. How to effectively utilize as much of existing abilities from source domains is a crucial issue in domain adaptation. 
In this paper, we propose an adversarial training procedure to train a Variational encoder-decoder based language generator via multiple adaptation steps. In this procedure, a model is first trained on a source domain data and then fine-tuned on a small set of target domain utterances under the guidance of two proposed critics.  
Experimental results show that the proposed method can effectively leverage the existing knowledge in the source domain to adapt to another related domain by using only a small amount of in-domain data. 
\end{abstract}

\section{Introduction}
\blfootnote{
    \hspace{-0.65cm}  
    This work is licensed under a Creative Commons 
    Attribution 4.0 International License.
    License details:
    \url{http://creativecommons.org/licenses/by/4.0/}.
}
Traditionally, Spoken Dialogue Systems are typically developed for various specific domains, including: finding a hotel, searching a restaurant \cite{wenthwsjy15}, or buying a tv, laptop \cite{wensclstm15}, flight reservations \cite{levin2000t}, etc. Such system are often requiring a well-defined ontology, which is essentially a data structured representation that the dialogue system can converse about. Statistical approaches to multi-domain in SDS system have shown promising results in how to reuse data in a domain-scalable framework efficiently \cite{young2013pomdp}. \newcite{mrkvsic2015multi} addressed the question of multi-domain in the SDS belief tracking by training a general model and adapting it to each domain. 

Recently, Recurrent Neural Networks (RNNs) based methods have shown improving results in tackling the domain adaptation issue \cite{chen2015recurrent,shi2015recurrent,wen2016multi,wentoward}. Such generators have also achieved promising results when providing such adequate annotated datasets \cite{wensclstm15,wenthwsjy15,tran-nguyen-tojo:2017:W17-55,tran-nguyen:2017:CoNLL,tran2017semantic}.
More recently, the development of the variational autoencoder (VAE) framework \cite{kingma2013auto,rezende2015variational} has paved the way for learning large-scale, directed latent variable models. This has brought considerable benefits to significant progress in natural language processing \cite{bowman2015generating,miao2016neural,purushotham2017variational,mnih2014neural}, dialogue system \cite{wen2017latent,serban2017hierarchical}.

This paper presents an adversarial training procedure to train a variational neural language generator via multiple adaptation steps, which enables the generator to learn more efficiently when in-domain data is in short supply. 
In summary,  we make the following contributions: (1) We propose a variational approach for an NLG problem which benefits the generator to adapt faster to new, unseen domain irrespective of scarce target resources; (2) We propose two critics in an adversarial training procedure, which can guide the generator to generate outputs that resemble the sentences drawn from the target domain; (3) We propose a unifying variational domain adaptation architecture which performs acceptably well in a new, unseen domain by using a limited amount of in-domain data; (4) We investigate the effectiveness of the proposed method in different scenarios, including ablation, domain adaptation, scratch, and unsupervised training with various amount of data.

\section{Related Work}
\label{sec:related_work}
Generally, Domain Adaptation involves two different types of datasets, one from a source domain and the other from a target domain. The source domain typically contains a sufficient amount of annotated data such that a model can be efficiently built, while there is often little or no labeled data in the target domain.
Domain adaptation for NLG have been less studied despite its important role in developing multi-domain SDS. \newcite{walker2001spot} proposed a SPoT-based generator to address domain adaptation problems. Subsequently, a system focused on tailoring user preferences \cite{walker2007individual}, and controlling user perceptions of linguistic style \cite{Mairesse:2011:CUP:2078087.2078089}. Moreover, a phrase-based statistical generator \cite{Mairesse:2010:PSL:1858681.1858838} using graphical models and active learning, and a multi-domain procedure \cite{wen2016multi} via data counterfeiting and discriminative training.

Neural variational framework for generative models of text have been studied longitudinally. 
\newcite{chung2015recurrent} proposed a recurrent latent variable model VRNN for sequential data by integrating latent random variables into hidden state of a RNN model. A hierarchical multi scale recurrent neural networks was proposed to learn both hierarchical and temporal representation \cite{chung2016hierarchical}. \newcite{2016arXiv160507869Z} introduced a variational neural machine translation that incorporated a continuous latent variable to model underlying semantics of sentence pairs. \newcite{bowman2015generating}  presented a variational autoencoder for unsupervised generative language model.

Adversarial adaptation methods have shown promising improvement in many machine learning applications despite the presence of domain shift or dataset bias, which reduce the difference between the training and test domain distributions, and thus improve generalization performance. 
\newcite{tzeng2017adversarial} proposed an improved unsupervised domain adaptation method to learn a discriminative mapping of target images to the source feature space by fooling a domain discriminator that tries to differentiate the encoded target images from source examples. 
We borrowed the idea of \cite{ganin2016domain}, where a domain-adversarial neural network are proposed  to learn features that are discriminative for the main learning task on the source domain, and indiscriminate with respect to the shift between domains.

\section{Variational Domain-Adaptation Neural Language Generator}
\label{sec:VDANLG}

Drawing inspiration from Variational autoencoder \cite{kingma2013auto} with assumption that there exists a continuous latent variable $z$ from a underlying semantic space of Dialogue Act (DA) and utterance pairs $(\textbf{d}, \textbf{y})$, we explicitly model the space together with variable $\textbf{d}$ to guide the generation process, \textit{i.e.}, $p(\textbf{y}|z, \textbf{d})$. With this assumption, the original conditional probability evolves to reformulate as follows:
\begin{equation}\label{eq:p-y-d-z}
p(\textbf{y}|\textbf{d})=\int_zp(\textbf{y},z|\textbf{d})\textbf{d}_z = \int_zp(\textbf{y}|z,\textbf{d})p(z|\textbf{d})\textbf{d}_z
\end{equation}
This latent variable enables us to model the underlying semantic space as a global signal for generation, in which the variational lower bound of variational generator can be formulated as follows:
\begin{equation}\label{eq:lowerbound}
\mathcal{L}_{VAE}(\theta, \phi, \textbf{d}, \textbf{y}) = -KL(q_\phi(z|\textbf{d}, \textbf{y}) || p_\theta(z|\textbf{d})) 	+ \mathbb{E}_{q_\phi(z|\textbf{d}, \textbf{y})}[\log p_\theta(\textbf{y}|z, \textbf{d})]
\end{equation}
where: $p_\theta(z|\textbf{d})$ is the prior model, $q_\phi(z|\textbf{d},\textbf{y})$ is the posterior approximator, and $p_\theta(\textbf{y}|z,\textbf{d})$ is the decoder with the guidance from global signal $z$, $KL(Q||P)$ is the Kullback-Leibler divergence between Q and P. 

\begin{figure*}[!ht] 
	\centering
    \includegraphics[width=0.85\textwidth, height=.5\textwidth]{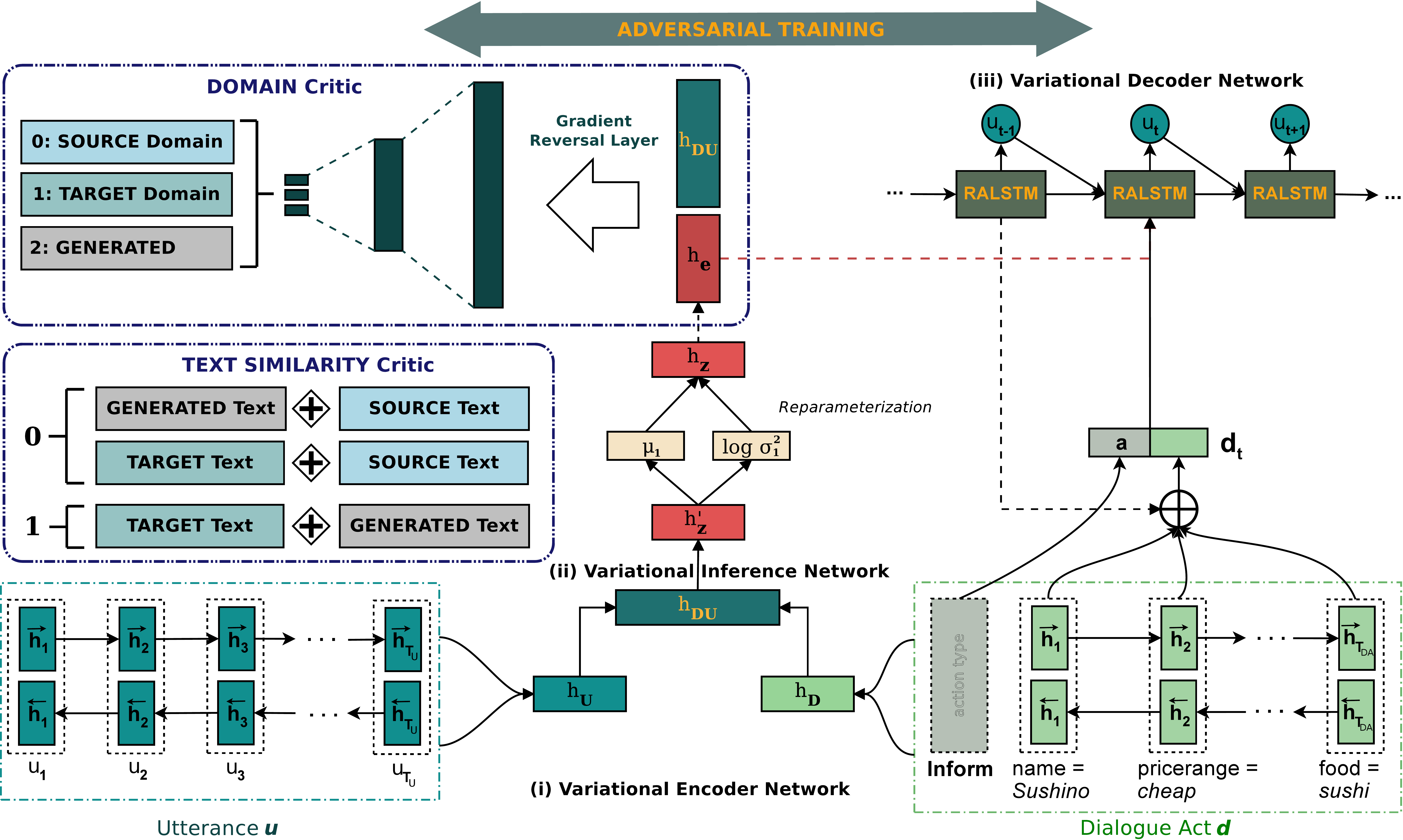}
    \vspace{-5pt}
    \caption{The VDANLG architecture which consists of two main components: the VRALSTM to generate the sentence and two Critics with an adversarial training procedure to guide the model in domain adaptation.}
    \label{fig:vnlg-model}\vspace{-5pt}
\end{figure*}

\subsection{Variational Neural Encoder}
The variational neural encoder aims at encoding a given input sequence $w_1, w_2, .., w_L$ into continuous vectors. In this work, we use a 1-layer, Bidirectional LSTM (BiLSTM) to encode the sequence embedding. The BiLSTM consists of forward and backward LSTMs, which read the sequence from left-to-right and right-to-left to produce both forward and backward sequence of hidden states ($\overrightarrow{\textbf{h}_{1}}, .., \overrightarrow{\textbf{h}_{L}}$), and  ($\overleftarrow{\textbf{h}_{1}}, .., \overleftarrow{\textbf{h}_{L}}$), respectively. We then obtain the sequence of encoded hidden states $\textbf{h}_\textbf{E}=(\textbf{h}_{1}, \textbf{h}_{2}, .., \textbf{h}_{L})$ where: $\textbf{h}_{i}=\overrightarrow{\textbf{h}_{i}} + \overleftarrow{\textbf{h}_{i}}$. 
We utilize this encoder to represent both the sequence of slot-value pairs $\{{\textbf{sv}}_i\}^{T_{DA}}_{i=1}$ in a given Dialogue Act, and the corresponding utterance $\{\textbf{y}_i\}^{T_Y}_{i=1}$ (see the red parts in Figure \ref{fig:vnlg-model}). We finally operate the mean-pooling over the BiLSTM hidden vectors to obtain the representation: $\textbf{h}_{D}=\frac{1}{T_{DA}} \sum_{i}^{T_{DA}}\textbf{h}_{i}, \textbf{h}_{Y}=\frac{1}{T_{Y}} \sum_{i}^{T_{Y}}\textbf{h}'_{i}$. The encoder, accordingly, produces both the DA representation vector which flows into the inferer and decoder, and the utterance representation which streams to the posterior approximator.

\subsection{Variational Neural Inferer}
In this section, we describe our approach to model both the prior $p_\theta(z|\textbf{d})$ and the posterior $q_\phi(z|\textbf{d},\textbf{y})$ by utilizing neural networks.

\subsection*{Neural Posterior Approximator}
Modeling the true posterior $p(z|\textbf{d},\textbf{y})$ is usually intractable. Traditional approach fails to capture the true posterior distribution of $z$ due to its oversimplified assumption when using the mean-field approaches. Following the work of \cite{kingma2013auto}, in this paper we employ neural network to approximate the posterior distribution of $z$ to simplify the posterior inference. We assume the approximation has the following form:
\begin{equation}\label{eq:approximation-form}
q_\phi(z|\textbf{d},\textbf{y}) = \mathcal{N}(z; \mu(f(\textbf{h}_D,\textbf{h}_Y)), \sigma^2(f(\textbf{h}_D, \textbf{h}_Y))\textbf{\textit{I}})
\end{equation}
where: the mean $\mu$ and standard variance $\sigma$ are the outputs of the neural network based on the representations of $\textbf{h}_D$ and $\textbf{h}_Y$. The function $f$ is a non-linear transformation that project the both DA and utterance representations onto the latent space:
\begin{equation}\label{eq:approximation-form}
\textbf{h}'_z = f(\textbf{h}_D, \textbf{h}_Y)=g(\textbf{W}_z[\textbf{h}_D;\textbf{h}_Y]+b_z)
\end{equation}
where: $\textbf{W}_z\in \mathbb{R}^{d_z \times (d_{\textbf{h}_D}+d_{\textbf{h}_Y})}$, $b_z \in \mathbb{R}^{d_z}$ are matrix and bias parameters respectively, $d_z$ is the dimensionality of the latent space, $g(.)$ is an elements-wise activation function which we set to be $Relu$ in our experiments.
In this latent space, we obtain the diagonal Gaussian distribution parameter $\mu$ and $\log\sigma^2$ through linear regression:
\begin{equation}\label{eq:posterior}
\mu = \textbf{W}_{\mu}\textbf{h}'_z + b_{\mu}, \log \sigma^2=\textbf{W}_{\sigma}\textbf{h}'_z+b_{\sigma}
\end{equation}
where: $\mu$, $\log\sigma^2$ are both $d_z$ dimension vectors.

\subsection*{Neural Prior Model}
We model the prior as follows:
\begin{equation}\label{eq:prior}
p_\theta(z|\textbf{d}) = \mathcal{N}(z; \mu'(\textbf{d}), \sigma'(\textbf{d})^2\textbf{\textit{I}})
\end{equation}
where: $\mu'$ and $\sigma'$ of the prior are neural models based on DA representation only, which are the same as those of the posterior $q_\phi(z|\textbf{d},\textbf{y})$ in Eq. \ref{eq:approximation-form} and Eq. \ref{eq:posterior}, except for the absence of $\textbf{h}_Y$. To acquire a representation of the latent variable $z$, we utilize the same technique as proposed in VAE \cite{kingma2013auto} and re-parameterize it as follows:\vspace{-5pt}
\begin{equation}\label{eq:reparameterization}
\textbf{h}_z = \mu + \sigma \odot \epsilon, \epsilon \sim \mathcal{N}(0, \textbf{\textit{I}})\vspace{-5pt}
\end{equation}

In addition, we set $\textbf{h}_z$ to be the mean of the prior $p_\theta(z|\textbf{d})$, \textit{i.e.}, $\mu'$, during decoding due to the absence of the utterance $y$. Intuitively, by parameterizing the hidden distribution this way, we can back-propagate the gradient to the parameters of the encoder and train the whole network with stochastic gradient descent. Note that the parameters for the prior and the posterior are independent of each other.

In order to integrate the latent variable $\textbf{h}_z$ into the decoder, we use a non-linear transformation to project it onto the output space for generation:\vspace{-5pt}
\begin{equation}\label{eq:reparameterization}
\textbf{h}_e = g(\textbf{W}_e\textbf{h}_z + b_e)\vspace{-5pt}
\end{equation}
where: $\textbf{h}_e \in \mathbb{R}^{d_e}$. It is important to notice that due to the sample noise $\epsilon$, the representation of $\textbf{h}_e$ is not fixed for the same input DA and model parameters. This benefits the model to learn to quickly adapt to a new domain (see Table \ref{tab:union-couterfeit}-(a) and Table \ref{tab:tab-performance}, \textit{sec.} 3).

\subsection{Variational Neural Decoder}
Given a DA $\textbf{d}$ and the latent variable $z$, the decoder calculates the probability over the generation $\textbf{y}$ as a joint probability of ordered conditionals:\vspace{-5pt}
\begin{equation}\label{eq:reparameterization}
p(\textbf{y}|z,\textbf{d})=\prod_{j=1}^{T_Y} p(\textbf{y}_t|\textbf{y}_{<t}, z, \textbf{d})\vspace{-5pt}
\end{equation}
where: $p(\textbf{y}_t|\textbf{y}_{<t}, z, \textbf{d}) = g'(RNN(\textbf{y}_{t}, \textbf{h}_{t-1}, \textbf{d}_{t})$
In this paper, we borrow the $\textbf{d}_t$ calculation and the computational RNN cell from \cite{tran-nguyen:2017:CoNLL} where RNN(.)=RALSTM(.) with a slightly modification in order to integrate the representation of latent variable, \textit{i.e.}, $\textbf{h}_e$, into the RALSTM cell, which is denoted by the bold dashed orange arrow in Figure \ref{fig:vnlg-model}-(iii). We modify the cell calculation as follows:
\begin{equation}\label{eq:lstm}
\begin{aligned}
\begin{pmatrix}
\textbf{i}_{t}
\\ \textbf{f}_{t}
\\ \textbf{o}_{t}
\\ \hat{\textbf{c}}_{t}
\end{pmatrix}
&=
\begin{pmatrix}\sigma
\\ \sigma
\\ \sigma
\\ \tanh
\end{pmatrix}\textbf{W}_{4d_h,4d_h}
\begin{pmatrix}
\textbf{h}_{e}
\\ \textbf{d}_{t}
\\ \textbf{h}_{t-1}
\\ \textbf{y}_{t}
\end{pmatrix}\\
\end{aligned}
\end{equation}
where: $\textbf{i}_{i}$, $\textbf{f}_{t}$, $\textbf{o}_{t}$ are input, forget and output gates respectively, $d_h$ is hidden layer size, $\textbf{W}_{4d_h,4d_h}$ is model parameter. 

The resulting Variational RALSTM (VRALSTM) model is demonstrated in Figure \ref{fig:vnlg-model}-(i), (ii), (iii), in which the latent variable can affect the hidden representation through the gates. This allows the model can indirectly take advantage of the underlying semantic information from the latent variable $z$. In addition, when the model learns to adapt to a new domain with unseen dialogue act, the semantic representation $\textbf{h}_e$ can help to guide the generation process (see \textit{sec.} \ref{subsec:distance} for details).

\subsection{Critics}
In this section, we introduce a \textit{text-similarity critic} and a \textit{domain critic} to guarantee, as much as possible, that the generated sentences resemble the sentences drawn from the target domain.
\subsection*{Text similarity critic}
To check the relevance between sentence pair in two domains and to encourage the model generating sentences in the style which is highly \textit{similar} to those in the target domain, we propose a Text Similarity Critic (SC) to classify $(\textbf{y}_{(1)}, \textbf{y}_{(2)})$ as 1-similar or 0-unsimilar text style. The model SC consists of two parts: a shared BiLSTM $\textbf{h}_Y$ with the Variational Neural Encoder to represent the $\textbf{y}_{(1)}$ sentence, and a second BiLSTM to encode the $\textbf{y}_{(2)}$ sentence. The SC model takes input as a pair $(\textbf{y}_{(1)}, \textbf{y}_{(2)})$ of ([\textit{target}], \textit{source}), ([\textit{target}], \textit{generated}), and ([\textit{generated}], \textit{source}). Note that we give priority to encoding the $\textbf{y}_{(1)}$ sentence in [.] using the shared BiLSTM, which guides the model to learn the sentence style from the target domain, and also contributes the target domain information into the global latent variables. We further utilize Siamese recurrent architectures \cite{neculoiu2016learning} for learning sentence similarity, in which the architecture allows us to learn useful representations with limited supervision.

\subsection*{Domain critic}
In consideration of the shift between domains, we introduce a Domain Critic (DC) to classify sentence as \textit{source}, \textit{target}, or \textit{generated} domain, respectively. 
Drawing inspiration from work of \cite{ganin2016domain}, we model DC with a gradient reversal layer and two standard feed-forward layers. It is important to notice that our DC model shares parameters with the Variational Neural Encoder and the Variational Neural Inferer. The DC model takes input as a pair of given DA and corresponding utterance to produce a concatenation of both its representation and its latent variable in the output space, which is then passed through a feed-forward layer and a 3-labels classifier. In addition, the gradient reversal layer, which multiplies the gradient by a specific negative value during back-propagation training, ensures that the feature distributions over the two domains are made similar, as indistinguishable as possible for the domain critic, hence resulting in the domain-invariant features. 

\section{Training Domain Adaptation Model}
\label{sec:training_da_model}
Given a training instance represented by a pair of DA and sentence $(\textbf{d}^{(i)}, \textbf{y}^{(i)})$ from the rich source domain $\mathcal{S}$ and the limited target domain $\mathcal{T}$, the task aims at finding a set of parameters $\Theta_{\mathcal{T}}$ that can perform acceptably well on the target domain.

\subsection{Training Critics}
We provide as following the training objective of SC and DC. For SC, the goal is to classify a sentence pair into $1$-\textit{similar} or $0$-\textit{unsimilar} textual style. This procedure can be formulated as a supervised classification training objective function: 
\begin{equation}\label{eq:sc-objective-function}
\left.
\begin{aligned}
&\mathcal{L}_s(\psi) = -\sum_{n=1}^N \log C_s(l^n_s|\textbf{y}_{(1)}^n, \textbf{y}_{(2)}^n, \psi), l^n_s=\left\{ 
	\begin{matrix*}[l]
    	1-similar    &  \textnormal{if } (\textbf{y}_{(1)}^n, \textbf{y}_{(2)}^n) \in \mathcal{P}_{sim}, \\
        0-unsimilar  &  \textnormal{if } (\textbf{y}_{(1)}^n, \textbf{y}_{(2)}^n) \in \mathcal{P}_{unsim},
	\end{matrix*}
	\right. \\
&
\mathcal{Y}_\mathcal{G} = \{\textbf{y} | \textbf{y} \sim \mathcal{G}(.|\textbf{d}_\mathcal{T},.) \}, \mathcal{P}_{sim} = \{\textbf{y}_{\mathcal{T}}^n , \textbf{y}^n_{\mathcal{Y}_\mathcal{G}}\}, 
\mathcal{P}_{unsim} = (\{\textbf{y}^n_{\mathcal{T}}, \textbf{y}^n_{\mathcal{S}}\},\{ \textbf{y}^n_{\mathcal{Y}_\mathcal{G}}, \textbf{y}^n_{\mathcal{S}}\})
\end{aligned}
\right.
\end{equation}
where: $N$ is number of sentences, $\psi$ is the model parameters of SC, $\mathcal{Y}_\mathcal{G}$ denotes sentences generated from the current generator $\mathcal{G}$ given target domain dialogue act $\textbf{d}_{\mathcal{T}}$. The scalar probability $C_s(1|\textbf{y}^n_{\mathcal{T}}, \textbf{y}^n_{\mathcal{Y}_\mathcal{G}})$ indicates how a generated sentence $\textbf{y}^n_{\mathcal{Y}_\mathcal{G}}$ is relevant to a target sentence $\textbf{y}^n_{\mathcal{T}}$. 

The DC critic aims at classifying a pair of DA-utterance into \textit{source}, \textit{target}, or \textit{generated} domain. This can also be formulated as a supervised classification training objective as follows:
\begin{equation}\label{eq:dc-objective-function}
\mathcal{L}_d(\varphi) = -\sum_{n=1}^N \log C_d(l^n_d|\textbf{d}^n, \textbf{y}^n, \varphi), l^n_d=\left\{ 
	\begin{matrix*}[l]
    	source    &  \textnormal{if } (\textbf{d}^n, \textbf{y}^n) \in (\mathcal{D}_\mathcal{S}, \mathcal{Y}_\mathcal{S}), 		\\
        target    &  \textnormal{if } (\textbf{d}^n, \textbf{y}^n) \in (\mathcal{D}_\mathcal{T}, \mathcal{Y}_\mathcal{T}), 		\\
        generated &  \textnormal{if } (\textbf{d}^n, \textbf{y}^n) \in (\mathcal{D}_\mathcal{T}, \mathcal{Y}_\mathcal{G}),
	\end{matrix*}
    \right.
\end{equation}
where: $\varphi$ is the model parameters of DC, ($\mathcal{D}_{S}, \mathcal{Y}_{S}$), ($\mathcal{D}_{T}, \mathcal{Y}_{T}$) are the DA-utterance pairs from source, target domain, respectively.
Note also that the scalar probability $C_d(target|\textbf{d}^n,\textbf{y}^n)$ indicates how likely the DA-utterance pair ($\textbf{d}^n, \textbf{y}^n$) is from the target domain.

\subsection{Training Variational Generator}
We utilize the Monte Carlo method to approximate the expectation over the posterior in Eq. \ref{eq:lowerbound}, \textit{i.e.}, $\mathbb{E}_{q_\phi(z|\textbf{d}, \textbf{y})}[.] \simeq \frac{1}{M}\sum_{m=1}^{M} \log p_\theta(\textbf{y}|\textbf{d}, \textbf{h}_z^{(m)})$ where: $M$ is the number of samples. In this study, the joint training objective for a training instance $(\textbf{d}, \textbf{y})$ is formulated as follows:
\begin{equation}\label{eq:g-objective-function}
\mathcal{L}(\theta, \phi) \simeq -KL(q_\phi(z|\textbf{d},\textbf{y})||p_\theta(z|\textbf{d}))
+ \frac{1}{M}\sum_{m=1}^{M} \sum_{t=1}^{T_{y}}\log p_\theta(\textbf{y}_t|\textbf{y}_{<t}, \textbf{d}, \textbf{h}_z^{(m)})
\end{equation}
where: $\textbf{h}_z^{(m)} = \mu + \sigma \odot \epsilon^{(m)}$, and $\epsilon^{(m)} \sim \mathcal{N}(0, \textbf{I})$. The first term is the KL divergence between two Gaussian distribution, and the second term is the approximation expectation. We simply set $M=1$ which degenerates the second term to the objective of conventional generator. Since the objective function in Eq. \ref{eq:g-objective-function} is differentiable, we can jointly optimize the parameter $\theta$ and variational parameter $\phi$ using standard gradient ascent techniques.

\subsection{Adversarial Training}
Our domain adaptation architecture is demonstrated in Figure \ref{fig:vnlg-model}, in which both generator $\mathcal{G}$ and critics $C_s$, and $C_d$ jointly train by pursuing competing goals as follows. Given a dialogue act $\textbf{d}_\mathcal{T}$ in the target domain, the generator generates $K$ sentences $\textbf{y}$'s. It would prefer a ``\textit{good}" generated sentence $\textbf{y}$ if the values of $C_d(target|\textbf{d}_\mathcal{T}, \textbf{y})$ and $C_s(1| \textbf{y}_\mathcal{T}, \textbf{y})$ are large. In contrast, the critics would prefer large values of $C_d(generated|\textbf{d}_\mathcal{T}, \textbf{y})$ and $C_s(1|\textbf{y}, \textbf{y}_\mathcal{S})$, which imply the small values of $C_d(target|\textbf{d}_\mathcal{T}, \textbf{y})$ and $C_s(1|\textbf{y}_\mathcal{T}, \textbf{y})$. We propose a domain-adversarial training procedure in order to iteratively updating the generator and critics as described in Algorithm \ref{algorithm:adv_training}. While the parameters of generator are optimized to minimize their loss in the training set, the parameters of the critics are optimized to minimize the error of text similarity, and to maximize the loss of domain classifier.

\begin{algorithm}
{\fontsize{10pt}{9pt}\selectfont
	\caption{Adversarial Training Procedure}
    \label{algorithm:adv_training}
    \SetAlgoLined
    \SetKwInput{Input}{Require}
    \SetKwInput{Output}{Output}
    \Input{generator $\mathcal{G}$, domain critic $C_d$, text similarity critic $C_s$, generated sentence $\mathcal{Y}_{\mathcal{G}} = \emptyset$;}
    \KwIn{ DA-utterance pairs of source $(\mathcal{D}_\mathcal{S}, \mathcal{Y}_{\mathcal{S}})$, target $(\mathcal{D}_\mathcal{T}, \mathcal{Y}_{\mathcal{T}})$;}
    Pretrain $\mathcal{G}$ on $(\mathcal{D}_\mathcal{S}, \mathcal{Y}_{\mathcal{S}})$ using VRALSTM;\\
    \While{$\Theta$ has not converged}{
    	\For{i = 0, .., $N_{\mathcal{T}}$}{   
        	Sample $(\textbf{d}_\mathcal{S}, \textbf{y}_\mathcal{S})$ from source domain;\\
        	($D_1$)-Compute $g_d = \triangledown_\varphi\mathcal{L}_d(\varphi)$ using Eq. \ref{eq:dc-objective-function} for $(\textbf{d}_\mathcal{S}, \textbf{y}_\mathcal{S})$ and $(\textbf{d}_\mathcal{T}, \textbf{y}_\mathcal{T})$;	\\
            ($D_2$)-Adam update of $\varphi$ for $C_d$ using $g_d$;\\
            ($G_1$)-Compute $g_\mathcal{G}=\{\triangledown_{\theta}\mathcal{L}(\theta, \phi),\triangledown_{\phi}\mathcal{L}(\theta, \phi)\}$ using Eq. \ref{eq:g-objective-function}\\
            ($G_2$)-Adam update of $\theta, \phi$ for $\mathcal{G}$ using $g_\mathcal{G}$\\
            ($S_1$)-Compute $g_s = \triangledown_\psi\mathcal{L}_s(\psi)$ using Eq. \ref{eq:sc-objective-function} for $(\textbf{y}_\mathcal{T}, \textbf{y}_\mathcal{S})$;	\\
            ($S_2$)-Adam update of $\psi$ for $C_s$ using $g_s$;\\
            $\mathcal{Y}_{\mathcal{G}} \leftarrow \{\textbf{y}_{\bar{k}}\}_{\bar{k}=1}^K$, where $\textbf{y}_{\bar{k}}\sim\mathcal{G}(.|\textbf{d}^{(i)}_\mathcal{T},.)$;\\
            Choose top $\textbf{\textit{k}}$ best sentences of $\mathcal{Y}_{\mathcal{G}}$; \\
            \For{j = 1,..,\textbf{k}}{
            	($D_1$), ($D_2$) steps for $C_d$ with $(\textbf{d}_\mathcal{T}, \textbf{y}^{(j)}_\mathcal{G})$;\\
                ($S_1$), ($S_2$) steps for $C_s$ with $(\textbf{y}^{(j)}_\mathcal{G}, \textbf{y}_\mathcal{S})$ and $(\textbf{y}_\mathcal{T},\textbf{y}^{(j)}_\mathcal{G})$;
            }
        }
    }
}
\end{algorithm}

Generally, the current generator $\mathcal{G}$ for each training iteration $i$ takes a \textit{target} dialogue act $\textbf{d}^{(i)}_\mathcal{T}$ as input to over-generate a set $\mathcal{Y}_\mathcal{G}$ of $K$ candidate sentences (step 11). We then choose top $\textbf{k}$ best sentences in the $\mathcal{Y}_\mathcal{G}$ set (step 12) after re-ranking to measure how ``\textit{good}" the generated sentences are by using the critics (steps 14-15). These ``\textit{good}" signals from the critics can guide the generator step by step to generate the outputs which resemble the sentences drawn from the target domain. Note that the re-ranking step is important for separating the ``\textit{correct}" sentences from the current generated outputs $\mathcal{Y}_\mathcal{G}$ by penalizing the generated sentences which have redundant or missing slots. 

\section{Experiments}
\label{sec:experiments}
We conducted experiments on the proposed models in different scenarios: \textit{\textbf{Adaptation}}, \textit{\textbf{Scratch}}, and \textit{\textbf{All}} using several model architectures, evaluation metrics, datasets \cite{wen2016multi}, and configurations (see Appendix A). 

\textbf{KL cost annealing strategy} \cite{bowman2015generating} encourages the model to encode meaningful representations into the latent vector $z$, in which we gradually anneal the KL term from $0$ to $1$. This helps our model to achieve solutions with non-zero KL term.

\textbf{Gradient reversal layer} \cite{ganin2016domain} leaves the input unchanged during forward propagation and reverses the gradient by multiplying it with a negative scalar $-\lambda_p$ during the backpropagation-based training. We set the domain adaptation parameter $\lambda_p$ which gradually increases, starting from $0$ to $1$, by using the following schedule for each training step $i$: $p = float(i)/num\_steps$, and $\lambda_p = \frac{2}{1+exp(-10 * p)}-1$ where: $num\_steps$ is a constant which is set to be $8600$, $p$ is training progress. This strategy allows the Domain critic to be less sensitive to noisy signal at the early training stages. 

\section{Results and Analysis}
\label{sec:results_analysis}

\subsection{Integrating Variational Inference}
We compare the original model RALSTM with its modification by integrating Variational Inference (VRALSTM) as demonstrated in Table \ref{tab:tab-scratch100} and Table \ref{tab:union-couterfeit}-(a). It clearly shows that the VRALSTM not only preserves the power of the original RALSTM on generation task since its performances are very competitive to those of RALSTM, but also provides a compelling evidence on adapting to a new, unseen domain when the target domain data is scarce, \textit{i.e.}, from $1$\% to $7$\%. 
Table \ref{tab:tab-performance}, \textit{sec.} 3 further shows the necessity of the integrating in which the VRALSTM achieved a significant improvement over the RALSTM in \textbf{\textit{Scratch}} scenario, and of the adversarial domain adaptation algorithm in which although both the RALSTM and VRALSTM model can perform well when providing sufficient in-domain training data (Table \ref{tab:tab-scratch100}), the performances are extremely impaired when training from \textbf{\textit{Scratch}} with only a limited data.
These indicate that the proposed variational method can learn the underlying semantic of DA-utterance pairs in the source domain via the representation of the latent variable $z$, from which when adapting to another domain, the models can leverage the existing knowledge to guide the generation process.

\begin{table*}[!ht]\vspace{-5pt}
\centering
\caption{Results when adapting models trained on (a) union, and (b) counterfeting dataset.}\vspace{-5pt}
\label{tab:union-couterfeit} 
\begin{tabularx}{\textwidth}{p{4.2cm}p{11cm}}
 \parbox[c]{4.2cm}{\includegraphics[width=4.2cm, height=3cm]{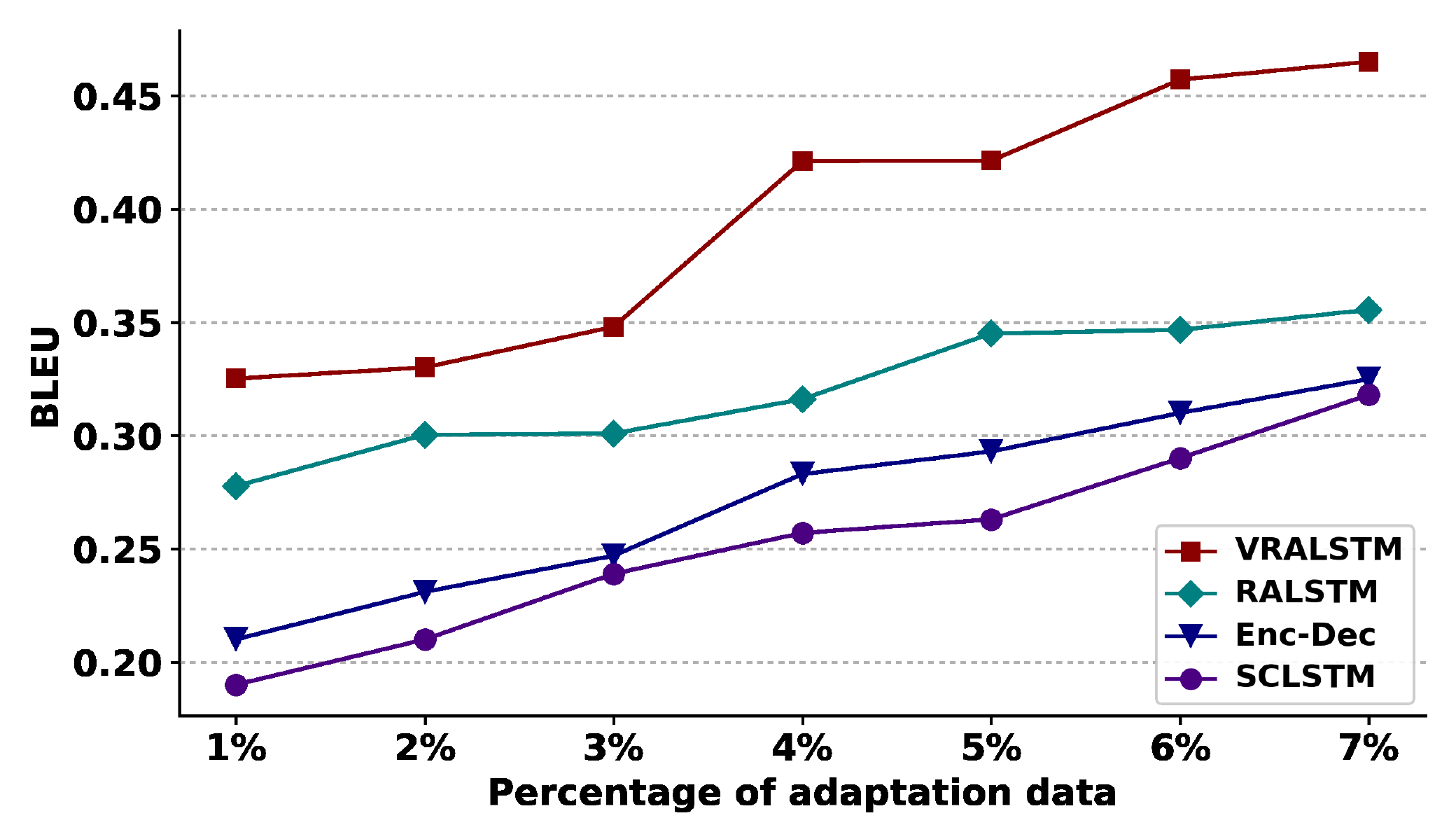}} 
 &
 \scalebox{0.65}{
\begin{tabular}{ccccccccc}
\hline 
\multirow{2}{*}{\diagbox{Source}{\textbf{Target}(\textit{Test})}} & \multicolumn{2}{c}{\textbf{R2H}(\textit{Hotel})} & \multicolumn{2}{c}{\textbf{H2R}(\textit{Restaurant})} & \multicolumn{2}{c}{\textbf{L2T}(\textit{Tv})} 		& \multicolumn{2}{c}{\textbf{T2L}(\textit{Laptop})} \\ \cline{2-9} 
							& BLEU 	  & ERR	  	& BLEU    & ERR     & BLEU    & ERR     & BLEU    & ERR     \\ \hline
Hotel$^{\sharp}$  			& - 	  & - 	    & 0.5931 & 12.50\%  & 0.4183  & 2.38\%  & 0.3426 & 13.02\% \\
Restaurant$^{\sharp}$   	& 0.6224  & 1.99\%  & - 	 & - 	    & 0.4211  & 2.74\%  & 0.3540 & 13.13\% \\
Tv$^{\sharp}$ 				& 0.6153  & 4.30\%  & 0.5835 & 14.49\%  & - 	  & - 	    & 0.3630 & 7.44\% \\
Laptop$^{\sharp}$ 			& 0.6042  & 5.22\%  & 0.5598 & 15.61\%  & 0.4268  & 1.05\%  & -      & - 	\\ \hline
\end{tabular}
}
    \\
(a) Result on \textbf{Laptop} when adapting models trained on [Restaurant+Hotel] data.
& 
(b) Results evaluated on (\textit{Test}) domains by \textbf{\textit{Unsupervised}} adapting VDANLG from Source domains using only \textbf{10\%} of the \textbf{Target} domain \textbf{Counterfeit X2Y}. \{\textbf{X},\textbf{Y}\}=\textbf{R} : Restaurant, \textbf{H} : Hotel, \textbf{T} : Tv, \textbf{L} : Laptop.
\end{tabularx}
\end{table*}

\vspace{-10pt}
\begin{table*}[!ht]\vspace{-5pt}
\centering
\caption{Results evaluated on \textbf{Target} domains by training models from scratch with \textbf{\textit{All}} in-domain data.}
\label{tab:tab-scratch100}\vspace{-5pt}
\scalebox{0.82}{
\begin{tabular}{ccccccccc}
\hline 
\multirow{2}{*}{\diagbox{Model}{\textbf{Target}}} & \multicolumn{2}{c}{\textbf{Hotel}} & \multicolumn{2}{c}{\textbf{Restaurant}} & \multicolumn{2}{c}{\textbf{Tv}} 		& \multicolumn{2}{c}{\textbf{Laptop}} \\ \cline{2-9} 
							& BLEU 	  & ERR 	& BLEU   & ERR     & BLEU    & ERR     & BLEU   & ERR     \\ \hline
HLSTM \cite{wenthwsjy15} 	& 0.8488  & 2.79\%  & 0.7436 & 0.85\%  & 0.5240  & 2.65\%  & 0.5130 & 1.15\% \\
SCLSTM \cite{wensclstm15} 	& 0.8469  & 3.12\%  & 0.7543 & 0.57\%  & 0.5235  & 2.41\%  & 0.5109 & 0.89\% \\
Enc-Dec \cite{wentoward}    & 0.8537  & 4.78\%  & 0.7358 & 2.98\%  & 0.5142  & 3.38\%  & 0.5101 & 4.24\% \\
RALSTM \cite{tran-nguyen:2017:CoNLL}  & 0.8911  & 0.48\%  & 0.7739 & 0.19\%  & 0.5376  & 0.65\% & 0.5222 & 0.49\% \\
VRALSTM (Ours)   			& 0.8851  & 0.57\%  & 0.7709 & 0.36\%  & 0.5356  & 0.73\%  & 0.5210 & 0.59\% \\ \hline
\end{tabular}
}
\vspace{-5pt}
\end{table*}

\begin{table*}[!ht]
\centering
\caption[Ablation studies' results comparison on scratch and adaptation training]{Ablation studies' results evaluated on \textbf{Target} domains by \textit{adaptation} training proposed models from Source domains using only \textit{10\%} amount of the \textbf{Target} domain data (\textit{sec.} 1, 2, 4, 5). The results were averaged over 5 randomly initialized networks.}
\label{tab:tab-performance}
\begin{adjustbox}{max width=0.85\textwidth}
\begin{tabular}{c|ccccccccc}
\hline 
&\multirow{2}{*}{\diagbox{Source}{\textbf{Target}}} & \multicolumn{2}{c}{\textbf{Hotel}} & \multicolumn{2}{c}{\textbf{Restaurant}} & \multicolumn{2}{c}{\textbf{Tv}} 		& \multicolumn{2}{c}{\textbf{Laptop}} \\ \cline{3-10} 
&						& BLEU 	  & ERR 	& BLEU   & ERR     & BLEU    & ERR     & BLEU   & ERR   \\ 

\hline 
\parbox[t]{2mm}{\multirow{6}{*}{\rotatebox[origin=c]{90}{\textbf{no Critics}}}}
&Hotel 		& - 	  & - 	    & 0.6814 & 11.62\% & 0.4968  & 12.19\% & 0.4915 & 3.26\% \\
&Restaurant   	& 0.7983  & 8.59\%  & - 	 & - 	   & 0.4805  & 13.70\% & 0.4829 & 9.58\% \\
&Tv 			& 0.7925  & 12.76\% & 0.6840 & 8.16\%  & - 	 	 & - 	   & 0.4997 & 4.79\% \\
&Laptop		& 0.7870  & 15.17\% & 0.6859 & 7.55\%  & 0.4953  & 18.60\% & - 		& - 	\\  
&{[R+H]} 		& - 	  & - 	    & - 	 & -       & 0.5019  & 7.43\%  & 0.4977 & 5.96\% \\
&{[L+T]} 		& 0.7935  & 11.71\% & 0.6927 & 6.49\%  & - 	 	 & - 	   & - 		& - 	 \\ 
\hline 
\parbox[t]{2mm}{\multirow{6}{*}{\rotatebox[origin=c]{90}{\textbf{+ DC + SC}}}}
&Hotel  		& - 	  & - 	    & 0.7131 & 2.53\%  & 0.5164  & 3.25\%  & 0.5007 & 1.68\% \\
&Restaurant  & 0.8217  & 3.95\%  & - 	 & - 	   & 0.5043  & 2.99\%  & 0.4931 & 2.77\% \\
&Tv 			& 0.8251  & 4.89\%  & 0.6971 & 4.62\%  & - 	 	 & - 	   & 0.5009 & 2.10\% \\
&Laptop 		& 0.8218  & 2.89\%  & 0.6926 & 2.87\%  & 0.5243  & 1.52\%  & -      & - 	\\ 
&{[R+H]} 	& - 	  & - 	    & - 	 & -       & 0.5197  & 2.58\%  & 0.5009 & 1.61\% \\
&{[L+T]} 	& 0.8252  & 2.87\%  & 0.7066 & 3.73\%  & - 	 	 & - 	   & - 		& - 	 \\ 
\hline
\parbox[t]{2mm}{\multirow{2}{*}{\rotatebox[origin=c]{90}{\textbf{scr10}}}}
&RALSTM					& 0.6855  & 22.53\% & 0.6003 & 17.65\% & 0.4009  & 22.37\% & 0.4475	& 24.47\% \\
&VRALSTM				& 0.7378  & 15.43\% & 0.6417 & 15.69\% & 0.4392  & 17.45\% & 0.4851	& 10.06\% \\
\hline
\parbox[t]{2mm}{\multirow{6}{*}{\rotatebox[origin=c]{90}{\textbf{+ DC only}}}}
&Hotel  		& - 	  & - 	    & 0.6823 & 4.97\%  & 0.4322  & 27.65\%  & 0.4389 & 26.31\% \\
&Restaurant  & 0.8031  & 6.71\%  & - 	 & - 	   & 0.4169  & 34.74\%  & 0.4245 & 26.71\% \\
&Tv 			& 0.7494  & 14.62\% & 0.6430 & 14.89\% & - 	 	 & - 	    & 0.5001 & 15.40\% \\
&Laptop 		& 0.7418  & 19.38\% & 0.6763 & 9.15\%  & 0.5114  & 10.07\%  & -      & - 	\\ 
&{[R+H]} 	& - 	  & - 	    & - 	 & -       & 0.4257  & 31.02\%  & 0.4331 & 31.26\% \\
&{[L+T]}	& 0.7658  & 8.96\%  & 0.6831 & 11.45\% & - 	 	 & - 	   	& - 	 & - 	 \\ 
\hline
\parbox[t]{2mm}{\multirow{6}{*}{\rotatebox[origin=c]{90}{\textbf{+ SC only}}}}
&Hotel  			& - 	  & - 	    & 0.6976 & 5.00\%  & 0.4896  & 9.50\%  & 0.4919 & 9.20\% \\
&Restaurant   	& 0.7960  & 4.24\%  & - 	 & - 	   & 0.4874  & 12.26\% & 0.4958 & 5.61\% \\
&Tv 				& 0.7779  & 10.75\% & 0.7134 & 5.59\%  & - 	 	 & - 	   & 0.4913 & 13.07\% \\
&Laptop 			& 0.7882  & 8.08\%  & 0.6903 & 11.56\% & 0.4963  & 7.71\%  & - 		& - 	\\  
&{[R+H]} 			& - 	  & - 	    & - 	 & -       & 0.4950  & 8.96\%  & 0.5002 & 5.56\% \\
&{[L+T]} 			& 0.7588  & 9.53\%  & 0.6940 & 10.52\% & - 	 	 & - 	   & - 		& - 	 \\ 
\hline 
\multicolumn{10}{l}{\textit{sec.} 3: Training RALSTM and VRALSTM models from \textit{scratch} using $10\%$ of \textbf{Target} domain data;}\\
\end{tabular}
 \end{adjustbox}\vspace{-15pt}
\end{table*}

\subsection{Ablation Studies}
The ablation studies (Table \ref{tab:tab-performance}, \textit{sec.} 1, 2) demonstrate the contribution of two Critics, in which the models were assessed with either no Critics or both or only one. It clearly sees that combining both Critics makes a substantial contribution to increasing the BLEU score and decreasing the slot error rate by a large margin in every dataset pairs. A comparison of model adapting from source Laptop domain between VRALSTM without Critics (Laptop$^{\flat}$) and VDANLG (Laptop$^\sharp$) evaluated on the target \textbf{Hotel} domain shows that the VDANLG not only has better performance with much higher the BLEU score, $82.18$ in comparison to $78.70$, but also significantly reduce the ERR, from $15.17$\% down to $2.89$\%. The trend is consistent across all the other domain pairs.
These stipulate the necessary Critics in effective learning to adapt to a new domain. 

Table \ref{tab:tab-performance}, \textit{sec.} 4 further demonstrates that using DC only (\textit{sec.} 4) brings a benefit of effectively utilizing similar slot-value pairs seen in the training data to \textit{closer} domain pairs such as: Hotel$\rightarrow$Restaurant ($68.23$ BLEU, $4.97$ ERR), Restaurant$\rightarrow$Hotel ($80.31$ BLEU, $6.71$ ERR), Laptop$\rightarrow$Tv ($51.14$ BLEU, $10.07$ ERR), and Tv$\rightarrow$Laptop ($50.01$ BLEU, $15.40$ ERR) pairs. Whereas it is inefficient for the \textit{longer} domain pairs since their performances are worse than those without Critics, or in some cases even worse than the VRALSTM in \textbf{\textit{scr10}} scenario, such as Restaurant$\rightarrow$Tv ($41.69$ BLEU, $34.74$ ERR), and the cases where Laptop to be a Target domain.
On the other hand, using only SC (\textit{sec.} 5) helps the models achieve better results since it is aware of the sentence style when adapting to the target domain.

\subsection{Distance of Dataset Pairs} \label{subsec:distance}
To better understand the effectiveness of the methods, we analyze the learning behavior of the proposed model between different dataset pairs. The datasets' order of difficulty was, from easiest to hardest: Hotel$\leftrightarrow$Restaurant$\leftrightarrow$Tv$\leftrightarrow$Laptop. On the one hand, it might be said that the \textit{longer} datasets' distance is, the more difficult of domain adaptation task becomes. This clearly shows in Table \ref{tab:tab-performance}, \textit{sec.} 1, at \textbf{Hotel} column where the adaptation ability gets worse regarding decreasing the BLEU score and increasing the ERR score alongside the order of Restaurant$\rightarrow$Tv$\rightarrow$Laptop datasets. On the other hand, the \textit{closer} the dataset pair is, the faster model can adapt. It can be expected that the model can better adapt to the target \textbf{Tv}/\textbf{Laptop} domain from source Laptop/Tv than those from source Restaurant, Hotel, and vice versa, the model can easier adapt to the target \textbf{Restaurant}/\textbf{Hotel} domain from source Hotel/Restaurant than those from Laptop, Tv. However, the above-mentioned is not always true that the proposed method can perform acceptably well from \textit{easy} source domains (Hotel, Restaurant) to the more \textit{difficult} target domains (Tv, Laptop) and vice versa (Table \ref{tab:tab-performance}, \textit{sec.} 1, 2). 

Table \ref{tab:tab-performance}, \textit{sec.} 2 further shows that the proposed method is able to leverage the out of domain knowledge since the adaptation models trained on union source dataset, such as [R+H] or [L+T], show better performances than those trained on individual source domain data. A specific example in Table \ref{tab:tab-performance}, \textit{sec.} 2 shows that the adaptation VDANLG model trained on the source union dataset of Laptop and Tv ([L+T]$^\sharp$) has better performance, at $82.52$ BLEU and $2.87$ ERR, than those models trained on the individual source dataset, such as Laptop$^\sharp$ ($82.18$ BLEU, $2.89$ ERR), and Tv$^\sharp$ ($82.51$ BLEU, $4.89$ ERR). Another example in Table \ref{tab:tab-performance}, \textit{sec.} 2 also shows that the adaptation VDANLG model trained on the source union dataset of Restaurant and Hotel ([R+H]$^\sharp$) has better results, at $51.97$ BLEU and $2.58$ ERR, than those models trained on the separate source dataset, such as Restaurant$^\sharp$ ($50.43$ BLEU, $2.99$ ERR), and Hotel$^\sharp$ ($51.64$ BLEU, $3.25$ ERR). The trend is mostly consistent across all other domain comparisons in different training scenarios.
All these demonstrate that the proposed model can learn global semantics that can be efficiently transferred into new domains.

\subsection{\textit{Adaptation} vs. \textit{All} Training Scenario}
It is interesting to compare \textit{\textbf{Adaptation}} (Table \ref{tab:tab-performance}, \textit{sec.} 2) with \textbf{\textit{All}} training scenario (Table \ref{tab:tab-scratch100}). The VDANLG model shows its considerable ability to shift to another domain with a limited of in-domain labels whose results are competitive to or in some cases better than the previous models trained on full labels of the \textbf{Target} domain. A specific comparison evaluated on the \textbf{Tv} domain where the VDANLG model trained on the source Laptop$^\sharp$ achieved better performance, at $52.43$ BLEU and $1.52$ ERR, than HLSTM ($52.40$, $2.65$), SCLSTM ($52.35$, $2.41$), and Enc-Dec ($51.42$, $3.38$). The VDANLG models, in many cases, also have lower of slot error rate ERR scores than the Enc-Dec model.
These indicate the stable strength of the VDANLG models in adapting to a new domain when the target domain data is scarce.

\begin{table*}[!ht]
\centering
\caption{Comparison of top \textbf{Laptop} responses generated for different scenarios by adaptation training VRALSTM (denoted by $\flat$) and VDANLG (denoted by ${\sharp}$) models from Source domains, and by training VRALSTM from \textit{\textbf{Scratch}}. Errors are marked in colors (\textcolor{red}{[missing]}, \textcolor{violet}{misplaced}, \textcolor{orange}{redundant}, \textcolor{teal}{wrong}, \textcolor{olive}{spelling mistake} information). \textcolor{blue}{[OK]} denotes successful generation. VDANLG$^\sharp$ = VRALSTM$^\flat$+SC+DC.}\vspace{-5pt}
\label{tab:lap-comparison}
\resizebox{1\textwidth}{!}{%
\begin{tabularx}{1.19\textwidth}{p{0.117\textwidth}p{1.03\textwidth}}
\textbf{Model} & \textbf{Generated Responses from Laptop Domain} \\ \hline 
\textbf{\textit{DA 1}} & compare(name=`\textit{tecra erebus 20}'; memory=`\textit{4 gb}'; isforbusinesscomputing=`\textit{true}'; name=`\textit{satellite heracles 45}'; memory=`\textit{2 gb}'; isforbusinesscomputing=`\textit{false}') \\
\textbf{\textit{Reference 1}} & compared to \textit{tecra erebus 20} which has a \textit{4 gb} memory and \textit{is for business computing} , \textit{satellite heracles 45} has a \textit{2 gb} memory and \textit{is not for business computing} . which one do you prefer \\ \hline

VRALSTM & which would be the \textit{tecra erebus 20} \textit{is a business computing} laptop with \textit{4 gb} of memory and is the \textcolor{orange}{SLOT\_NAME} , and \textcolor{violet}{is not for business computing} . \textcolor{red}{[satellite heracles 45]}\textcolor{red}{[2 gb]}\\ \hline 

Hotel$^{\flat}$ & \textit{the tecra erebus 20} \textit{is used for business computing} . the \textit{satellite heracles 45} has \textcolor{violet}{4 gb} of memory and a \textcolor{teal}{SLOT\_BATTERY} battery life \textcolor{orange}{for business computing} . which one do you want \\

Restaurant$^{\flat}$ & the \textit{tecra erebus 20} \textit{is for business computing} . the \textit{satellite heracles 45} which has \textcolor{violet}{4 gb} of memory and \textit{is not for business computing} . which one do you want \textcolor{red}{[2 gb]}\\

Tv$^{\flat}$ & \textit{the tecra erebus 20} has \textit{4 gb} of memory and \textcolor{violet}{is not for business computing} . which one do you prefer \textcolor{red}{[is for business computing]}\textcolor{red}{[satellite heracles 45]}\textcolor{red}{[2 gb]} \\

{[R+H]}$^{\flat}$ & the \textit{tecra erebus 20} \textcolor{violet}{is not for business computing} . \textcolor{olive}{which one do you want a business computing} . which one do you prefer \textcolor{red}{[4 gb]}\textcolor{red}{[is for business computing]}\textcolor{red}{[satellite heracles 45]}\textcolor{red}{[2 gb]} \\ \hline

Hotel$^{\sharp}$ & the \textit{tecra erebus 20} has a \textit{4 gb} memory , that \textit{is for business computing} . the \textit{satellite heracles 45} with \textit{2 gb} of memory and \textit{is not for business computing} . which one do you want \textcolor{blue}{[OK]}\\

Restaurant$^{\sharp}$ & the \textit{tecra erebus 20} has a \textit{4 gb} memory , and \textit{is for business computing} . the \textit{satellite heracles 45} \textit{is not for business computing} . which one do you want to know more \textcolor{red}{[2 gb]} \\

Tv$^{\sharp}$ & the \textit{tecra erebus 20} \textit{is a business computing} . the \textit{satellite heracles 45} has a \textcolor{violet}{4 gb} memory and \textit{is not for business computing} . which one do you prefer \textcolor{red}{[2 gb]}\\

{[R+H]}$^{\sharp}$ & the \textit{tecra erebus 20} \textit{is for business computing} , has a \textcolor{violet}{2 gb} of memory. the \textit{satellite heracles 45} has \textcolor{violet}{4 gb} of memory , \textit{is not for business computing}. which one do you want \\ \hline
\end{tabularx} %
}
\vspace{-10pt}
\end{table*}

\subsection{Unsupervised Domain Adaptation}
We further examine the effectiveness of the proposed methods by training the VDANLG models on target \textbf{\textit{Counterfeit}} datasets \cite{wen2016multi}. The promising results are shown in Table \ref{tab:union-couterfeit}-(b), despite the fact that the models were instead adaptation trained on the \textbf{Counterfeit} datasets, or in other words, were indirectly trained on the (\textit{Test}) domains. However, the proposed models still showed positive signs in remarkably reducing the slot error rate ERR in the cases of \textit{Hotel} and \textit{Tv} be the (\textit{Test}) domains. Surprisingly, even the source domains (Hotel$^\sharp$/Restaurant$^\sharp$) are far from the (\textit{Test}) domain \textit{Tv}, and the \textbf{Target} domain \textbf{Counterfeit L2T} is also very different to the source domains, the model can still acceptably adapt well since its BLEU scores on (\textit{Test}) Tv domain reached to ($41.83$/$42.11$), and it also produced a very low of ERR scores ($2.38$/$2.74$). 
This phenomenon will be further investigated in the unsupervised scenario in the future work.

\subsection{Comparison on Generated Outputs}
On the one hand, the VRALSTM models (trained from \textit{\textbf{Scratch}} or trained adapting model from Source domains) produce the outputs with a diverse range of error types, including \textcolor{red}{missing}, \textcolor{violet}{misplaced}, \textcolor{orange}{redundant}, \textcolor{teal}{wrong} slots, or even \textcolor{olive}{spelling mistake} information, leading to a very high of the slot error rate ERR score. Specifically, the VRALSTM from \textit{\textbf{Scratch}} tends to make repeated slots and also many of the missing slots in the generated outputs since the training data may inadequate for the model to generally handle the unseen dialog acts. Whereas the VRALSTM models without Critics adapting trained from Source domains (denoted by $\flat$ in Table \ref{tab:lap-comparison} and Appendix B. Table \ref{tab:laptv-comparison}) tend to generate the outputs with fewer error types than the model from \textit{\textbf{Scratch}} because the VRALSTM$^\flat$ models may capture the overlap slots of both source and target domain during adaptation training. 

On the other hand, under the guidance of the Critics (SC and DC) in an adversarial training procedure, the VDANLG model (denoted by ${\sharp}$) can effectively leverage the existing knowledge of the source domains to better adapt to the target domains. The VDANLG models can generate the outputs in style of the target domain with much fewer the error types compared with the two above models. Moreover, the VDANLG models seem to produce satisfactory utterances with more correct generated slots. For example, a sample outputted by the [R+H]$^\sharp$ in Table \ref{tab:lap-comparison}-example 1 contains all the required slots with only a \textcolor{violet}{misplaced} information of two slots \textcolor{violet}{\textit{2 gb}} and \textcolor{violet}{\textit{4 gb}}, while the generated output produced by Hotel$^\sharp$ is a successful generation. 
Another samples in Appendix B. Table \ref{tab:laptv-comparison} generated by the Hotel$^\sharp$, Tv$^\sharp$, [R+H]$^\sharp$ (in DA 2) and Laptop$^\sharp$ (DA 3) models are all fulfilled responses. 
An analysis of the generated responses in Table \ref{tab:laptv-comparison}-example 2 illustrates that the VDANLG models seem to generate a concise response since the models show a tendency to form some potential slots into a concise phrase, \textit{i.e.}, ``SLOT\_NAME SLOT\_TYPE". For example, the VDANLG models tend to concisely response as ``the \underline{\textit{portege phosphorus 43}} \underline{\textit{laptop}} ..." instead of ``the \underline{\textit{portege phosphorus 43}} is a \underline{\textit{laptop}} ...".
All these above demonstrate that the VDANLG models have ability to produce better results with a much lower of the slot error rate ERR score.

\section{Conclusion and Future Work}
\label{sec:conclusion_future_work}
We have presented an integrating of a variational generator and two Critics in an adversarial training algorithm to examine the model ability in domain adaptation task. 
Experiments show that the proposed models can perform acceptably well in a new, unseen domain by using a limited amount of in-domain data. 
The ablation studies also demonstrate that the variational generator contributes to effectively learn the underlying semantic of DA-utterance pairs, while the Critics show its important role of guiding the model to adapt to a new domain. The proposed models further show a positive sign in unsupervised domain adaptation, which would be a worthwhile study in the future. 

\section*{Acknowledgements}
This work was supported by the JST CREST Grant Number JPMJCR1513, the JSPS KAKENHI Grant number 15K16048 and the SIS project.

\bibliography{coling2018}
\bibliographystyle{acl} 

\subsection*{Appendix A. Experimental Setups}
We followed the configurations for the RALSTM model from work of \cite{tran-nguyen:2017:CoNLL}, in which the hidden layer size and beam width were set to be $80$ and $10$, respectively, and the generators were trained with a $70\%$ of keep dropout rate. We performed $5$ runs with different random initialization of the network, and the training process is terminated by using early stopping. We then selected a model that yields the highest BLEU score \cite{papineni2002bleu} on the validation set. For the variational inference, we set the latent variable size to be $16$. We used Adam optimizer with the learning rate is initially set to be $0.001$, and after $3$ epochs for the Union dataset and $5$ epochs for the single dataset the learning rate is decayed every epoch using an exponential rate of $0.95$. The decoding phase is mostly similar to those in work of \cite{tran-nguyen:2017:CoNLL}.

\subsection*{Appendix B. Generated Utterances}
Table \ref{tab:laptv-comparison} presents top responses generated for different scenarios from Laptop and TV domains. 

\begin{table*}[!ht]
\centering
\caption{Comparison of top responses generated for different scenarios by adaptation training VRALSTM (denoted by $\flat$) and VDANLG (denoted by ${\sharp}$) models from Source domains, and by training VRALSTM from \textit{\textbf{Scratch}}. Errors are marked in colors (\textcolor{red}{[missing]}, \textcolor{violet}{misplaced}, \textcolor{orange}{redundant}, \textcolor{teal}{wrong}, \textcolor{olive}{spelling mistake} information). \textcolor{blue}{[OK]} denotes successful generation. VDANLG$^\sharp$ = VRALSTM$^\flat$+SC+DC.}
\label{tab:laptv-comparison}
\resizebox{\textwidth}{!}{%
\begin{tabularx}{1.19\textwidth}{p{0.117\textwidth}p{1.03\textwidth}}
\hline
\textbf{Model} & \textbf{Generated Responses from Laptop Domain} \\ \hline 

\textbf{\textit{DA 2}} & inform(name=\textit{`portege phosphorus 43'}; type=\textit{`laptop'}; design=\textit{`black magnesium chassis with brushed metallic finish , matt black keyboard'}; isforbusinesscomputing=\textit{`false'}; drive=\textit{`320 gb'}) \\
\textbf{\textit{Reference 2}} & the \textit{portege phosphorus 43} \textit{laptop} has a \textit{320 gb} drive , \textit{is not for business computing} and \textit{has a black magnesium chassis with brushed metallic finish , matt black keyboard} \\ 
\hline
VRALSTM  & the \textit{portege phosphorus 43} is a \textit{laptop} with a \textit{320 gb} drive and \textit{has a black magnesium chassis with brushed metallic finish , matt black keyboard} . \textcolor{red}{[is not for business computing]}
\\ 
\hline 
Hotel$^{\flat}$ & the \textit{portege phosphorus 43} is a \textit{laptop} has a \textit{320 gb} drive , \textit{is not for business computing} . \textcolor{orange}{it is not for business computing} , it has a design of \textit{black magnesium chassis with brushed metallic finish , matt black keyboard}\\

Restaurant$^{\flat}$ & the \textit{portege phosphorus 43} is a \textit{laptop} with a \textit{320 gb} drive , has a design of \textit{black magnesium chassis with brushed metallic finish , matt black keyboard} . \textcolor{red}{[is not for business computing]}\\

Tv$^{\flat}$ & the \textit{portege phosphorus 43} is a \textit{laptop} with a \textit{black magnesium chassis with brushed metallic finish , matt black keyboard} . it is not for business computing \textcolor{red}{[320 gb]}\\

{[R+H]}$^{\flat}$ & the \textit{portege phosphorus 43} is a \textit{laptop} with a \textit{black magnesium chassis with brushed metallic finish , matt black keyboard} \textcolor{red}{[is not used for business computing]} \textcolor{red}{[320 gb]} 
\\ \hline

Hotel$^{\sharp}$ & the \textit{portege phosphorus 43} \textit{laptop} has a \textit{320 gb} drive , has a \textit{black magnesium chassis with brushed metallic finish , matt black keyboard} design and \textit{is not for business computing} \textcolor{blue}{[OK]}
\\
Restaurant$^{\sharp}$ & \textit{the portege phosphorus 43 laptop} has a \textit{320 gb} drive , \textcolor{teal}{it is for business computing} , it has a design of \textit{black magnesium chassis with brushed metallic finish , matt black keyboard}
\\
Tv$^{\sharp}$ & \textit{the portege phosphorus 43 laptop} has a \textit{320 gb} drive and a design of \textit{black magnesium chassis with brushed metallic finish , matt black keyboard} . it \textit{is not for business computing} \textcolor{blue}{[OK]}
 \\
{[R+H]}$^{\sharp}$ & the \textit{portege phosphorus 43 laptop} has a \textit{320 gb} drive , and \textit{is not for business computing} . it has a \textit{black magnesium chassis with brushed metallic finish , matt black keyboard} \textcolor{blue}{[OK]}
\\ \\ \hline

\textbf{Model} & \textbf{Generated Responses from TV Domain} \\ \hline 
\textbf{\textit{DA 3}} & compare(name=`\textit{crios 69}'; ecorating=`\textit{a++}'; powerconsumption=`\textit{44 watt}'; name=`\textit{dinlas 61}'; ecorating=`\textit{a+}'; powerconsumption=`\textit{62 watt}') \\
\textbf{\textit{Reference 3}} & compared to \textit{crios 69} which is in the \textit{a++} eco rating and has \textit{44 watt} power consumption , \textit{dinlas 61} is in the \textit{a+} eco rating and has \textit{62 watt} power consumption . which one do you prefer ? \\ \hline
VRALSTM & the \textit{crios 69} is the \textcolor{violet}{dinlas 61} is the \textcolor{orange}{SLOT\_NAME} is the \textcolor{orange}{SLOT\_NAME} is the \textcolor{orange}{SLOT\_NAME} is the \textcolor{orange}{SLOT\_NAME} is the \textcolor{orange}{SLOT\_NAME} is the \textcolor{orange}{SLOT\_NAME} is the \textcolor{orange}{SLOT\_NAME} . it has an \textcolor{violet}{a++} eco rating \textcolor{red}{[44 watt]}\textcolor{red}{[a+]}\textcolor{red}{[62 watt]}
\\ \hline 
Hotel$^{\flat}$ & the \textit{crios 69} has a \textit{44 watt} power consumption , whereas the \textit{dinlas 61} has \textit{62 watt} power consumption , whereas the \textcolor{orange}{SLOT\_NAME} has \textcolor{orange}{SLOT\_POWERCONSUMPTION} power consumption and has an \textcolor{violet}{a++} eco rating
 \textcolor{red}{[a+]} \\
 
Restaurant$^{\flat}$ & the \textit{crios 69} has a \textit{a++} eco rating , \textit{44 watt} power consumption , and an \textcolor{violet}{a+} eco rating and \textcolor{violet}{62 watt} power consumption \textcolor{red}{[dinlas 61]} \\

Laptop$^{\flat}$ & the \textit{crios 69} has \textcolor{teal}{SLOT\_HDMIPORT} hdmi port -s , the \textit{dinlas 61} has \textcolor{violet}{a++} eco rating and \textcolor{violet}{44 watt} power consumption \textcolor{red}{[62 watt]}\textcolor{red}{[a+]} \\

{[R+H]}$^{\flat}$ & the \textit{crios 69} is in the \textcolor{teal}{SLOT\_FAMILY} product family with a++ eco rating ? \textcolor{red}{[44 watt]}\textcolor{red}{[dinlas 61]}\textcolor{red}{[62 watt]}\textcolor{red}{[a+]} \\ \hline

Hotel$^{\sharp}$ & the \textit{crios 69} has an \textit{a++} eco rating and \textit{44 watt} power consumption and a \textcolor{violet}{62 watt} power consumption \textcolor{red}{[dinlas 61]}\textcolor{red}{[a+]} \\

Restaurant$^{\sharp}$ & the \textit{crios 69} has \textit{44 watt} power consumption of \textit{a++} and has an \textcolor{violet}{a+} eco rating and \textcolor{violet}{62 watt} power consumption \textcolor{red}{[dinlas 61]} \\

Laptop$^{\sharp}$ & the \textit{crios 69} has an \textit{a++} eco rating and \textit{44 watt} power consumption , whereas the \textit{dinlas 61} has \textit{62 watt} power consumption and \textit{a+} eco rating . \textcolor{blue}{[OK]} \\

{[R+H]}$^{\sharp}$ & the \textit{crios 69} has \textit{44 watt} power consumption , and an \textit{a++} eco rating and the \textit{dinlas 61} has a \textit{62 watt} power consumption . \textcolor{red}{[a+]} \\ \hline
\end{tabularx} %
}
\end{table*}

\end{document}